\documentclass[10pt,twocolumn,letterpaper]{article}

\usepackage{cvpr}
\usepackage{times}
\usepackage{url}            
\usepackage{booktabs}       
\usepackage{amsfonts}       
\usepackage{nicefrac}       
\usepackage{microtype}      

\usepackage{siunitx}
\usepackage{epsfig}
\usepackage{graphicx}
\usepackage{amsmath}
\usepackage{amssymb}
\usepackage{enumitem}
\usepackage{threeparttable}
\usepackage{makecell}
\usepackage{bbding}
\usepackage[percent]{overpic}
\usepackage{multirow}

\usepackage[breaklinks=true,bookmarks=false]{hyperref}

\cvprfinalcopy 

\begin{document}

\title{Learning Spatial Fusion for Single-Shot Object Detection}

\author{Songtao Liu\\
Beihang University\\
{\tt\small liusongtao@buaa.edu.cn}
\and
Di Huang\\
Beihang University\\
{\tt\small dhuang@buaa.edu.cn}
\and
Yunhong Wang\\
Beihang University\\
{\tt\small yhwang@buaa.edu.cn}
}

\maketitle

\begin{abstract}
	Pyramidal feature representation is the common practice to address the challenge of scale variation in object detection. However, the inconsistency across different feature scales is a primary limitation for the single-shot detectors based on feature pyramid. In this work, we propose a novel and data driven strategy for pyramidal feature fusion, referred to as adaptively spatial feature fusion (ASFF). It learns the way to spatially filter conflictive information to suppress the inconsistency, thus improving the scale-invariance of features, and introduces nearly free inference overhead. With the ASFF strategy and a solid baseline of YOLOv3, we achieve the best speed-accuracy trade-off on the MS COCO dataset, reporting 38.1\% AP at 60 FPS, 42.4\% AP at 45 FPS and 43.9\% AP at 29 FPS. The code is available at \url{https://github.com/ruinmessi/ASFF}.
\end{abstract}

\section{Introduction}

Object detection is one of the most fundamental components in various downstream vision tasks. In recent years, the performance of object detectors has been remarkably improved thanks to the rapid development of deep convolutional neural networks (CNNs) \cite{alexnet,vgg,resnet,inceptionv1} and well-annotated datasets \cite{Pascal-voc,ms-coco}. However, handling multiple objects across a wide range of scales still remains a challenging problem. To achieve scale invariance, recent state-of-the-art detectors construct feature pyramids or multi-level feature towers \cite{ssd,FPN,yolov3,mask-rcnn,focal-loss}.

The Single Shot Detector (SSD) \cite{ssd} is one of the first attempts to generate convolutional pyramidal feature representations for object detection. It reuses the multi-scale feature maps from different layers computed in the forward pass to predict objects of various sizes. However, this bottom-up pathway suffers from low accuracies on small instances as the shallow-layer feature maps contain insufficient semantic information. To address the disadvantage of SSD, Feature Pyramid Network (FPN) \cite{FPN} sequentially combines two adjacent layers in feature hierarchy in the backbone model with a top-down pathway and lateral connections. The low-resolution, semantically strong features are up-sampled and combined with high-resolution, semantically weak features to build a feature pyramid that shares rich semantics at all levels. FPN and other similar top-down structures \cite{dssd,ron,stairnet,refinedet,yolov3} are simple and effective, but they still leave much room for improvement. Indeed, many recent models \cite{panet,weaving,dla,kong-eccv,libra} with advanced cross-scale connections show accuracy gains through strengthening feature fusion. Besides the manually designed fusion structures, NAS-FPN \cite{nas-fpn} applies Neural Architecture Search (NAS) techniques to pursue a better architecture, producing significant improvements upon many backbones. 

\begin{figure}
	\centering
	\includegraphics[width=0.99\linewidth]{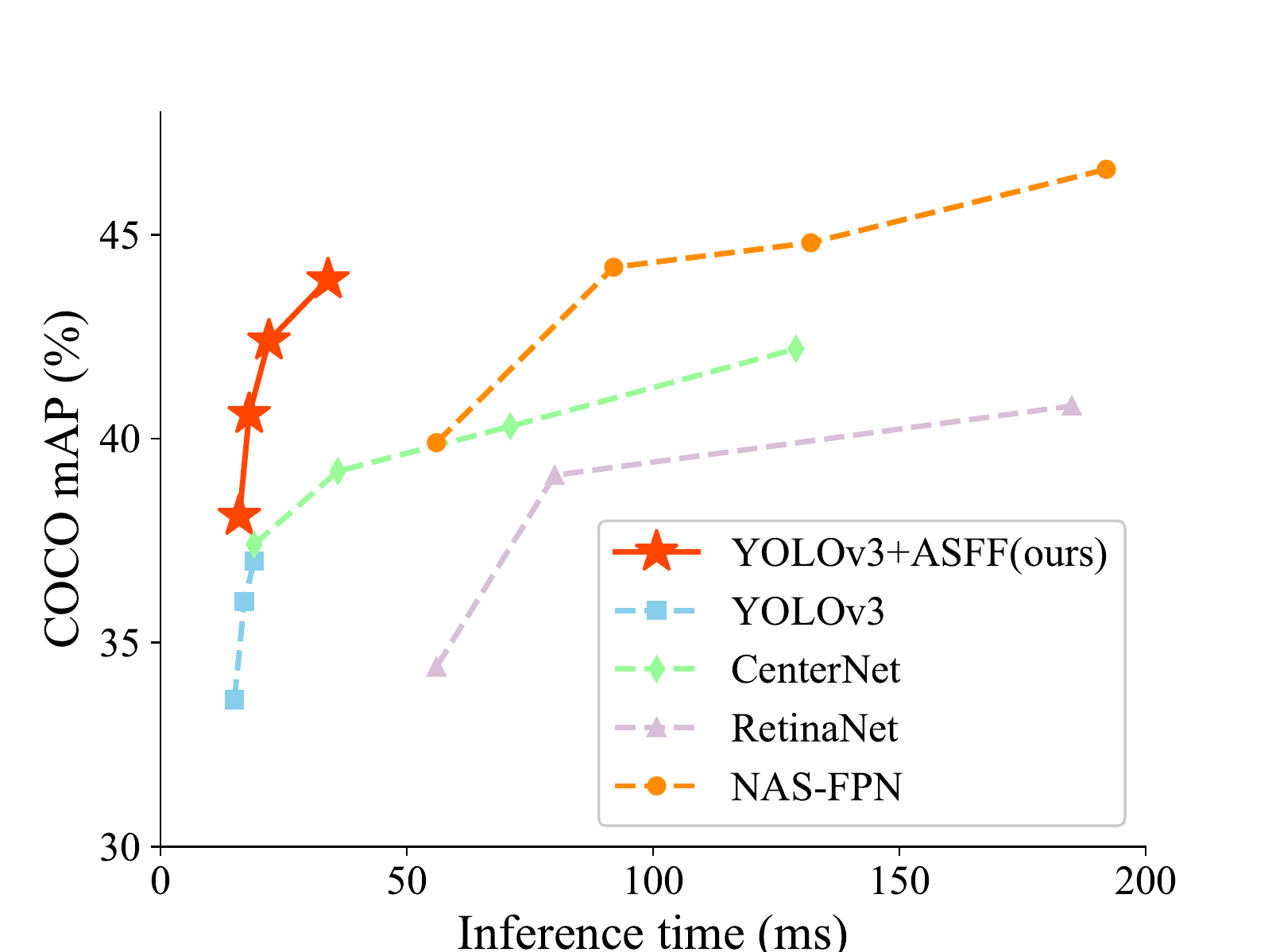}
	\vspace{0.1cm}
	\caption{Speed-accuracy trade-off on COCO test-dev for real-time detectors. The proposed ASFF helps YOLOv3 outperform a range of state-of-the-art algorithms.}
	\label{fig:speed}
\end{figure}

Although these advanced studies deliver more powerful feature pyramids, they still leave room for scale-invariant prediction. Some evidences are given by SNIP \cite{snip,sniper}, which adopts a scale normalization method that selectively trains and infers the objects of appropriate sizes in each image scale of the multi-scale image pyramids, achieving further improvements on the results of pyramidal feature based detectors with multi-scale testing. However, image pyramid solutions sharply increase the inference time, which makes them not applicable to real-world applications.

Meanwhile, compared to image pyramids, one main drawback of feature pyramids is the inconsistency across different scales, in particular for single-shot detectors. Specifically, when detecting objects with feature pyramids, a heuristic-guided feature selection is adopted: large instances are typically associated with upper feature maps and small instances with lower feature maps. When an object is assigned and treated as positive in the feature maps at a certain level, the corresponding areas in the feature maps of other levels are viewed as background. Therefore, if an image contains both small and large objects, the conflict among features at different levels tends to occupy the major part of the feature pyramid. This inconsistency interferes gradient computation during training and downgrades the effectiveness of feature pyramids. Some models adopt several tentative strategies to deal with this problem. \cite{guide,FSAF} set the corresponding areas of feature maps at adjacent levels as ignore regions (\emph{i.e.} zero gradients), but this alleviation may increase inferior predictions at the the adjacent levels of features. TridentNet \cite{trident} creates multiple scale-specific branches with different receptive fields for scale-aware training and inference. It breaks away from feature pyramids to avoid inconsistency, but also misses reusing its higher-resolution maps, limiting the accuracy of small instances.

In this paper, we propose a novel and effective approach, named adaptively spatial feature fusion (ASFF), to address the inconsistency in feature pyramids of single-shot detectors. The proposed approach enables the network to directly learn how to spatially filter features at other levels so that only useful information is kept for combination. For the features at a certain level, features of other levels are first integrated and resized into the same resolution and then trained to find the optimal fusion. At each spatial location, features at different levels are fused adaptively, \emph{i.e.}, some features may be filter out as they carry contradictory information at this location and some may dominate with more discriminative clues. ASFF offers several advantages: (1) as the operation of searching the optimal fusion is differential, it can be conveniently learned in back-propagation; (2) it is agnostic to the backbone model and it is applied to single-shot detectors that have a feature pyramid structure; and (3) its implementation is simple and the increased computational cost is marginal.

Experiments on the COCO \cite{ms-coco} benchmark confirm the effectiveness of our method. We first adopt the recent advanced training tricks \cite{bag} and anchor-guiding pipeline \cite{guide} to provide a solid baseline for YOLOv3 \cite{yolov3} (\emph{i.e.}, 38.8\% mAP with 50 FPS). We then employ ASFF to further improve this enhanced YOLOv3 and another strong single-stage detector, RetinaNet \cite{focal-loss}, equipped with different backbones by a large margin, while keeping computational cost under control. Especially, we boost the YOLOv3 baseline to 42.4\% mAP with 45 FPS and 43.9\% mAP with 29 FPS, which is a state-of-the-art speed and accuracy trade-off among all the existing detectors on COCO.

\section{Related Work}

\begin{figure*}[thbp]
	\begin{center}
		\includegraphics[width=0.90\linewidth]{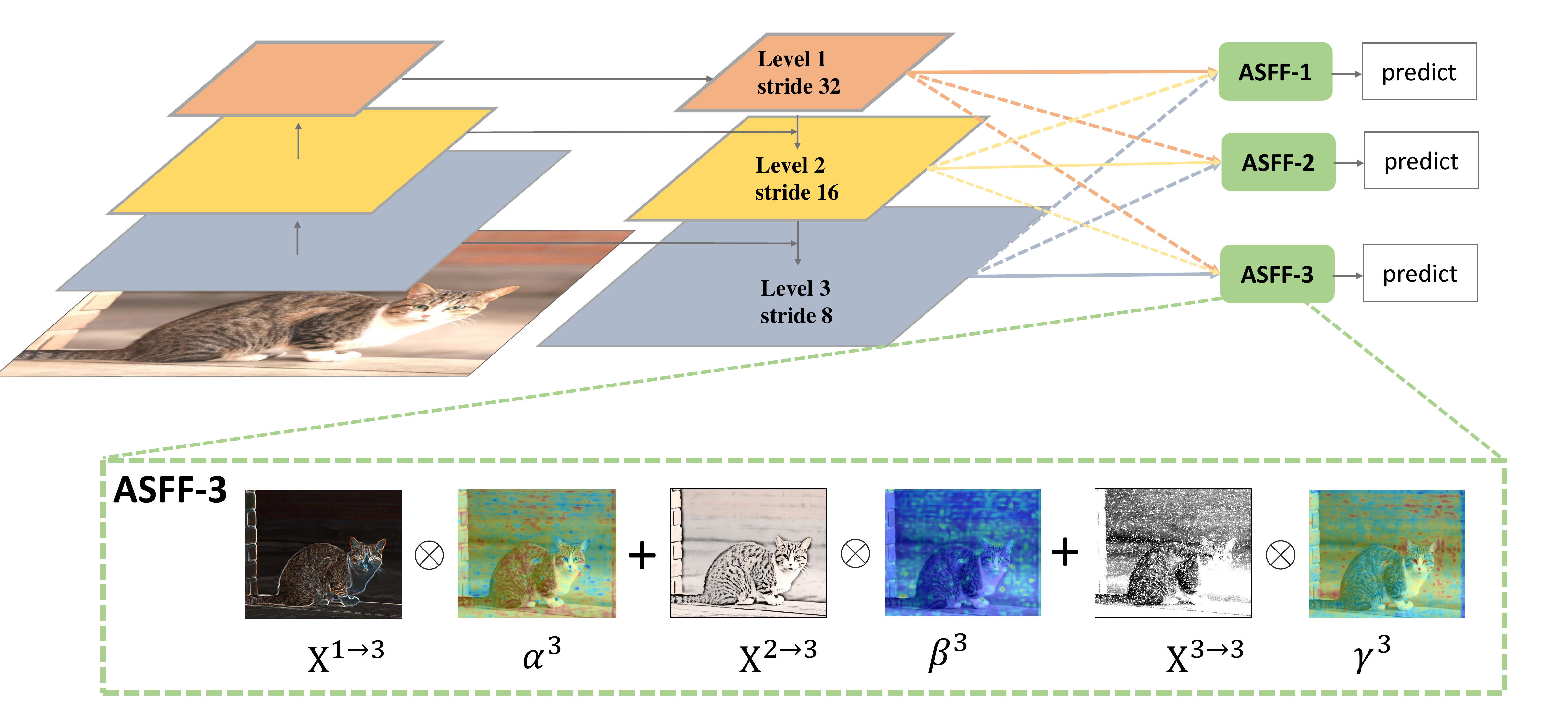}
	\end{center}
	\caption{Illustration of the adaptively spatial feature fusion mechanism. For each level, the features of all the other levels are resized to the same shape and spatially fused according to the learned weight maps.}
	\label{fig:frame}
\end{figure*}

Feature pyramid representations or multi-level feature towers are the basis of solutions of multi-scale processing in recent object detectors. SSD \cite{ssd} is one of the first attempts to predict class scores and bounding boxes from multiple feature scales in a bottom-up manner. FPN \cite{FPN} builds feature pyramid by sequentially combining two adjacent level of features with top-down pathway and lateral connections. Such connections effectively enhance feature representations and the rich semantics from depp and low-resolution features are shared at all levels.

Following FPN, many other models with similar top-down structures \cite{dssd,ron,stairnet,refinedet,yolov3}appear, which achieve substantial improvements for object detection. Recently, more advanced investigations have attempted to ameliorate such multi-scale feature representations. For instance, PANet \cite{panet} proposes an additional bottom-up pathway based on FPN to increase the low-level information in deep layers. Chen \emph{et al.} \cite{weaving} build a pyramid based on SSD that weaves features across different level of feature layers. DLA \cite{dla} introduces iterative deep aggregation and hierarchical deep aggregation structures to better fuse semantic and spatial information. Kim \emph{et al.} \cite{pfp} show a parallel feature pyramid by adopting spatial pyramid pooling and widening the network. Zhu \emph{et al.} \cite{FSAF} present a feature selective anchor-free module to dynamically choose the most suitable feature level for each instance. Kong \emph{et al.} \cite{kong-eccv} aggregate feature maps at all scales to a specific scale and then produce features at each scale by a global attention operation on the combined features. Libra R-CNN \cite{libra} also integrates features at all levels to generate more balanced semantical features. In addition to manually designing the fusion structure, NAS-FPN \cite{nas-fpn} applies the Neural Architecture Search algorithm to seek a more powerful fusion architecture, delivering the best single-shot detector.

In spite of competitive scores, those feature pyramid based methods still suffer from the inconsistency across different scales, which limits the further performance gain. To address this, \cite{guide,FSAF} set the corresponding regions of adjacent levels as ignore regions (\emph{i.e.} zero gradients), but the relaxation in the adjacent levels tends to cause more inferior predictions as false positives. TridentNet \cite{trident} drops out the structure of feature pyramids and creates multiple scale-specific branches with different receptive fields to adopt scale-aware training and inferencing, but the performance of small instances may suffer from the missing of its higher-resolution maps. 

The proposed ASFF approach alleviates this issue by learning connections among different feature maps. Actually, this idea is not new within the domain of computer vision. \cite{gated-2} adopts element-wise product in two adjacent feature maps to form a gate unit in a top-down manner for dense label prediction. The element-wise product operation reduces the categorical ambiguity from shallow layers and highlights the discriminability from deeper layers. This gating mechanism succeeds in semantic segmentation. However, the task of dense labeling does not need heuristic-guided feature selection required in object detection, since the features at all levels predict the same label map at different scales. It thus does not reduce spatial contradiction in object detection. \cite{gated-1} proposes a sigmoid gating unit in the skip connection between convolutional and deconvolutional layers of features at each single level for visual counting. It optimizes the flow of information within the feature maps at the same level, but does not deal with the inconsistency in feature pyramids. ACNet \cite{acnet} employs a flexible way to switch global and local inference in processing the feature representations by adaptively determining the connection status among the feature nodes from CNNs, classical multi-layer perceptron and non-local network. In contrast to them, ASFF adaptively learns the import degrees for different levels of features on each location to avoid spatial contradiction.

\section{Method}

In this section, we instantiate our adaptively spatial feature fusion (ASFF) approach by showing how it works on the single-shot detectors with feature pyramids, such as SSD \cite{ssd}, RetinaNet \cite{focal-loss}, and YOLOv3 \cite{yolov3}. Taking YOLOv3 as an example, we apply ASFF to it and demonstrate the resulting detector in the following steps. First, we push YOLOv3 to a baseline, much stronger than the origin \cite{yolov3}, by adopting the recent advanced training tricks \cite{bag} and anchor-free pipeline \cite{guide,FSAF}. Then, we present the formulation of ASFF and give a qualitative analysis of the consistency property of the pyramid feature fusion and ASFF. Finally, we display the details of training, testing, and implementing the models.

\subsection{Strong Baseline}

We take the YOLOv3 \cite{yolov3} framework because it is simple and efficient. In YOLOv3, there are two main components: an efficient backbone (DarkNet-53) and a feature pyramid network of three levels. A recent work \cite{bag} significantly improves the performance of YOLOv3 without modifying network architectures and bringing extra inference cost. Moreover, a number of studies \cite{guide,FSAF,fcos,foveabox} indicate that the anchor-free pipeline contributes to considerably better performance with simpler designs. To better demonstrate the effectiveness of our proposed ASFF approach, we build a baseline, much stronger than the origin \cite{yolov3}, based on these advanced techniques.

Following \cite{bag}, we introduce a bag of tricks in the training process, such as the mixup algorithm \cite{mixup}, the cosine \cite{cosine} learning rate schedule, and the synchronized batch normalization technique \cite{megdet}. Besides those tricks, we further add an anchor-free branch to run jointly with anchor-based ones as \cite{FSAF} does and exploit the anchor guiding mechanism proposed by \cite{guide} to refine the results. Moreover, an extra Intersection over Union (IoU) loss function \cite{iouloss} is employed on the original smooth L1 loss for better bounding box regression. More details can be found in the supplemental material. 

With these advanced techniques mentioned above, we achieve 38.8\% mAP on the COCO \cite{ms-coco} 2017 \emph{val} set at a speed of 50 FPS (on Tesla V100), improving the original YOLOv3-608 baseline (33.0\% mAP with 52 FPS \cite{yolov3}) by a large margin without heavy computational cost in inference.

\subsection{Adaptively Spatial Feature Fusion}
\label{ASFF}

Different from the former approaches that integrate multi-level features using element-wise sum or concatenation, our key idea is to adaptively learn the spatial weight of fusion for feature maps at each scale.
The pipeline is shown in Figure \ref{fig:frame}, and it consists of two steps: identically rescaling and adaptively fusing.

\paragraph{Feature Resizing.} We denote the features of the resolution at level $l$ ($l \in \{1,2,3\}$ for YOLOv3) as $\mathbf{x}^l$. For level $l$, we resize the features $\mathbf{x}^n$ at the other level $n\ (n\neq l)$ to the same shape as that of $\mathbf{x}^l$. Because the features at three levels in YOLOv3 have different resolutions as well as different numbers of channels, we accordingly modify the up-sampling and down-sampling strategies for each scale. For up-sampling, we first apply a $1\times1$ convolution layer to compress the number of channels of features to that in level $l$, and then upscale the resolutions respectively with interpolation. For down-sampling with $1/2$ ratio, we simply use a $3\times3$ convolution layer with a stride of 2 to modify the number of channels and the resolution simultaneously. For the scale ratio of $1/4$, we add a 2-stride max pooling layer before the 2-stride convolution.

\paragraph{Adaptive Fusion.} Let $\mathbf{x}_{ij}^{n\rightarrow l}$ denote the feature vector at the position $(i,j)$ on the feature maps resized from level $n$ to level $l$. We propose to fuse the features at the corresponding level $l$ as follows:
\begin{equation}
\label{eq:1}
\mathbf{y}_{ij}^l = \alpha^l_{ij} \cdot \mathbf{x}_{ij}^{1\rightarrow l} + \beta^l_{ij} \cdot \mathbf{x}_{ij}^{2\rightarrow l} +\gamma^l_{ij} \cdot \mathbf{x}_{ij}^{3\rightarrow l},
\end{equation}
where $\mathbf{y}_{ij}^l$ implies the $(i,j)$-th vector of the output feature maps $\mathbf{y}^l$ among channels. $\alpha^l_{ij}$, $\beta^l_{ij}$ and $\gamma^l_{ij}$ refer to the spatial importance weights for the feature maps at three different levels to level $l$, which are adaptively learned by the network. Note that $\alpha^l_{ij}$, $\beta^l_{ij}$ and $\gamma^l_{ij}$ can be simple scalar variables, which are shared across all the channels. Inspired by \cite{acnet}, we force $\alpha^l_{ij}+\beta^l_{ij}+\gamma^l_{ij}=1$ and $\alpha^l_{ij},\beta^l_{ij},\gamma^l_{ij} \in [0,1]$, and define
\begin{small}\begin{equation}
	\alpha^l_{ij} = \frac{e^{\lambda^l_{\alpha_{ij}}}}{e^{\lambda^l_{\alpha_{ij}}} + e^{\lambda^l_{\beta_{ij}
		}} + e^{\lambda^l_{\gamma_{ij}}}}.
	\end{equation}\end{small}
Here $\alpha^l_{ij}$, $\beta^l_{ij}$ and $\gamma^l_{ij}$ are defined by using the softmax function with $\lambda^l_{\alpha_{ij}}$, $\lambda^l_{\beta_{ij}}$ and $\lambda^l_{\gamma_{ij}}$ as control parameters respectively. We use $1\times1$ convolution layers to compute the weight scalar maps $\mathbf{\lambda}^l_\alpha$, $\mathbf{\lambda}^l_\beta$ and $\mathbf{\lambda}^l_\gamma$ from $\mathbf{x}^{1\rightarrow l}$, $\mathbf{x}^{2\rightarrow l}$ and $\mathbf{x}^{3\rightarrow l}$ respectively, and they can thus be learned through standard back-propagation.

With this method, the features at all the levels are adaptively aggregated at each scale. The outputs $\{\mathbf{y}^1,\mathbf{y}^2,\mathbf{y}^3\}$ are used for object detection following the same pipeline of YOLOv3.

\subsection{Consistency Property}
In this section, we analyze the consistency property of the proposed ASFF approach and the other alternatives of feature fusion. Without loss of generality, we focus on the gradient at a certain position $(i,j)$ of the unresized feature maps at level 1 $\mathbf{x}^1$ in YOLOv3. Following the chain rule, the gradient is computed as:
\begin{equation}
\label{eq:3}
\begin{aligned}
\frac{\partial \mathcal{L}}{\partial \mathbf{x}_{ij}^1} =&
\frac{\partial \mathbf{y}_{ij}^1}{\partial \mathbf{x}_{ij}^1} \cdot 
\frac{\partial \mathcal{L}}{\partial \mathbf{y}_{ij}^1}        + 
\frac{\partial \mathbf{x}_{ij}^{1\rightarrow 2}}{\partial \mathbf{x}_{ij}^1} \cdot
\frac{\partial \mathbf{y}_{ij}^2}{\partial \mathbf{x}_{ij}^{1\rightarrow 2}} \cdot 
\frac{\partial \mathcal{L}}{\partial \mathbf{y}_{ij}^2}       \\
& +
\frac{\partial \mathbf{x}_{ij}^{1\rightarrow 3}}{\partial \mathbf{x}_{ij}^1} \cdot
\frac{\partial \mathbf{y}_{ij}^3}{\partial \mathbf{x}_{ij}^{1\rightarrow 3}} \cdot 
\frac{\partial \mathcal{L}}{\partial \mathbf{y}_{ij}^3}
\end{aligned}
\end{equation} 

It is worth to note that feature resizing usually uses interpolation for up-sampling and pooling for down-sampling. We thus assume that $\frac{\partial \mathbf{x}_{ij}^{1\rightarrow l}}{\partial \mathbf{x}_{ij}^1}\approx 1$ for simplicity. Then Eq. (\ref{eq:3}) can be written as:
\begin{equation}
\label{eq:4}
\frac{\partial \mathcal{L}}{\partial \mathbf{x}_{ij}^1} =
\frac{\partial \mathbf{y}_{ij}^1}{\partial \mathbf{x}_{ij}^1} \cdot 
\frac{\partial \mathcal{L}}{\partial \mathbf{y}_{ij}^1}        + 
\frac{\partial \mathbf{y}_{ij}^2}{\partial \mathbf{x}_{ij}^{1\rightarrow 2}}  \cdot 
\frac{\partial \mathcal{L}}{\partial \mathbf{y}_{ij}^2}        +
\frac{\partial \mathbf{y}_{ij}^3}{\partial \mathbf{x}_{ij}^{1\rightarrow 3}}  \cdot 
\frac{\partial \mathcal{L}}{\partial \mathbf{y}_{ij}^3}
\end{equation}

For the two common fusion operations used in RetinaNet \cite{focal-loss}, YOLOv3 \cite{yolov3} and other pyramidal feature based detectors (\emph{i.e.} element-wise sum and concatenation), we can further simplify the equation to the following with $\frac{\partial \mathbf{y}_{ij}^1}{\partial \mathbf{x}_{ij}^1} =1$ and $\frac{\partial \mathbf{y}_{ij}^l}{\partial \mathbf{x}_{ij}^{1\rightarrow l}}=1$:
\begin{equation}
\label{eq:5}
\frac{\partial \mathcal{L}}{\partial \mathbf{x}_{ij}^1} =
\frac{\partial \mathcal{L}}{\partial \mathbf{y}_{ij}^1}        +  
\frac{\partial \mathcal{L}}{\partial \mathbf{y}_{ij}^2}        +
\frac{\partial \mathcal{L}}{\partial \mathbf{y}_{ij}^3}
\end{equation}

Suppose position $(i,j)$ at level 1 is assigned as the center of an object according to a certain scale matching mechanism and $\frac{\partial \mathcal{L}}{\partial \mathbf{y}_{ij}^1}$ is the gradient from  the positive sample. As the corresponding positions are viewed as background in the other levels, $\frac{\partial \mathcal{L}}{\partial \mathbf{y}_{ij}^2}$ and $\frac{\partial \mathcal{L}}{\partial \mathbf{y}_{ij}^3}$ are the gradients from negative samples. This inconsistency disturbs the gradient of $\frac{\partial \mathcal{L}}{\partial \mathbf{x}_{ij}^1}$ and downgrades the training efficiency of the original feature maps $\mathbf{x}^1$. 

One typical way to deal with this problem is to set the corresponding positions of the other levels as ignore regions (\emph{i.e.} $\frac{\partial \mathcal{L}}{\partial \mathbf{y}_{ij}^2}=\frac{\partial \mathcal{L}}{\partial \mathbf{y}_{ij}^3}=0$) \cite{guide,FSAF}. However, although the conflict in $\mathbf{x}_{ij}^1$ is eliminated, the relaxation in $\mathbf{y}_{ij}^2$ and $\mathbf{y}_{ij}^3$ tends to cause more inferior predictions as false positives at the suboptimal levels. 

For ASFF, it is straightforward to calculate the gradient from Eq. (\ref{eq:1}) and Eq. (\ref{eq:4}) as follows:
\begin{equation}
\label{eq:6}
\frac{\partial \mathcal{L}}{\partial \mathbf{x}_{ij}^1} =
\alpha_{ij}^1 \cdot 
\frac{\partial \mathcal{L}}{\partial \mathbf{y}_{ij}^1}        + 
\alpha_{ij}^2  \cdot 
\frac{\partial \mathcal{L}}{\partial \mathbf{y}_{ij}^2}        +
\alpha_{ij}^3  \cdot 
\frac{\partial \mathcal{L}}{\partial \mathbf{y}_{ij}^3},
\end{equation}
where $\alpha^1_{ij},\alpha^2_{ij},\alpha^3_{ij} \in [0,1]$. With these three coefficients, the inconsistency of gradient can be harmonized if $\alpha^2_{ij} \rightarrow 0$ and $\alpha^3_{ij} \rightarrow 0$. Since the fusion parameters can be learned by the standard back-propagation algorithm, a well-tuned training process can yield such effective coefficients (see some qualitative results in Figure \ref{fig:weight1} and Figure \ref{fig:weight2}). Meanwhile, the supervision information of the background in $\frac{\partial \mathcal{L}}{\partial \mathbf{y}_{ij}^2}$ and $\frac{\partial \mathcal{L}}{\partial \mathbf{y}_{ij}^2}$ is kept, avoiding generating more false positives.   

\subsection{Training, Inference, and Implementation}
\paragraph{Training.} Let $\Theta$ denote the set of network parameters (\emph{e.g.}, the weights of convolution filters) and $\Phi = \{\mathbf{\lambda}^l_\alpha, \mathbf{\lambda}^l_\beta,\mathbf{\lambda}^l_\gamma |\ l = 1,2,3 \}$ be the set of fusion parameters that control the spatial fusion of each scale. We jointly optimize the two sets of parameters by minimizing a loss function $\mathcal{L}(\Theta, \Phi)$, where $\mathcal{L}$ is the original YOLOv3 objective function plus the IoU regression loss \cite{iouloss} for both anchor shape prediction and bounding box regression. Following \cite{bag}, we apply mixup on the classification pretraining of DarkNet53, and all the new convolution layers are employed with the MSRA weight initialization method~\cite{MSRA}. To reduce the risk of overfitting and improve generalization of network predictions, we follow the approach of random shapes training as in YOLOv3 \cite{yolov3}. More specifically, a mini-batch of $N$ training images is resized to $N\times 3 \times H \times W$, where $H = W$ is randomly picked in $\{320, 352, 384, 416, 448, 480, 512, 544, 576, 608\}$.

\paragraph{Inference.} During inference, the detection header at each level first predicts the shape of anchors, and then conducts classification and box regression following the same pipeline as that in YOLOv3 \cite{yolov3}. Next, non-maximum suppression (NMS) with the threshold at 0.6 is applied to each class separately. For simplicity and fair comparison against other counterparts, we do not use the advanced testing tricks such as Soft-NMS \cite{soft_nms} or test-time image augmentations.

\paragraph{Implementation.} We implement the modified YOLOv3 as well as ASFF using the existing PyTorch v1.0.1 framework with CUDA 10.0 and CUDNN v7.1. The entire network is trained with stochastic gradient descent (SGD) on 4 GPUs (NVDIA Tesla V100) with 16 images per GPU. All models are trained for 300 epochs with the first 4 epochs of warmup and the cosine learning rate schedule \cite{cosine} from 0.001 to 0.00001. The weight decay is 0.0005 and the momentum is 0.9. We also follow the implementation of \cite{bag} to turn off mixup augmentation for the last 30 epochs.

\section{Experiments}
\begin{table*}[htbp]
	\begin{center}
		\scalebox{.8}{
			\resizebox{\textwidth}{!}{
				\begin{threeparttable}
					\begin{tabular}{cccclllllll}
						\hline
						YOLOv3 @608 & BoF \cite{bag} & GA \cite{guide} & IoU \cite{iouloss} & AP   & $AP_{50}$ & $AP_{70}$ & $AP_S$    & $AP_M$    & $AP_L$   & FPS \\ \hline
						\checkmark &     &    &     & 33.0 & 57.9 & 34.4 & 18.3 & 35.4 & 41.9 & 52\\
						\checkmark &  \checkmark  &    &     & 37.2 & 57.9 & 40.0 & 23.4 & 41.9 & 49.0 &52 \\
						\checkmark &  \checkmark   &   \checkmark &      & 38.2 & 58.3 & 40.8 & 23.9 & 42.6 & 49.9 &50\\
						\checkmark &  \checkmark    &   &   \checkmark   & 37.6 & 58.7 & 40.2 & 23.8 & 42.7 & 48.9 &52\\
						\checkmark &  \checkmark    &   \checkmark  &   \checkmark   & 38.8 & 58.3 & 43.0 & 24.6 & 42.9 & 51.6 &50\\ \hline
					\end{tabular}
		\end{threeparttable}}}
	\end{center}\
	\caption{Effect of each component on the baseline. Results in terms of AP (\%) and FPS are reported on COCO \emph{val}-2017.}
	\label{table:baseline}
\end{table*}

\begin{table}[htbp]
	\begin{center}
		\scalebox{.47}{
			\resizebox{\textwidth}{!}{
				\begin{threeparttable}
					\begin{tabular}{l|l|lll}
						\hline
						Method                                                                 & Ignore area     & AP   & AP50 & AP70 \\ \hline
						\multirow{4}{*}{\begin{tabular}[c]{@{}l@{}}YOLOv3\\ @608\end{tabular}}
						& baseline ($\epsilon_{ignore}$=0)   & 38.8 & 58.3 & 43.0 \\
						& center location & 38.8 & 58.3 & 43.1 \\
						& $\epsilon_{ignore}$=0.2           & 39.1 & 59.0 & 43.4 \\
						& $\epsilon_{ignore}$=0.5           & 37.5 & 57.3 & 40.5 \\ \hline
					\end{tabular}
		\end{threeparttable}}}
	\end{center}
	\caption{Improvement by the adjacent ignoring strategy. APs (\%) are reported on COCO \emph{val}-2017.}
	\label{table:ignore}
\end{table}

\begin{table}[htbp]
	\begin{center}
		\scalebox{.47}{
			\resizebox{\textwidth}{!}{
				\begin{threeparttable}
					\begin{tabular}{llllllll}
						\hline
						YOLOv3 @608 & AP   & $AP_{50}$ & $AP_{70}$ & $AP_S$    & $AP_M$    & $AP_L$   & FPS \\ \hline
						baseline        & 38.8 & 58.3 & 43.0 & 24.6 & 42.9 & 51.6 & 50\\
						baseline + concat        & 39.5 & 59.1 & 43.7 & 25.6 & 43.8 & 50.3 & 42\\
						baseline + sum        & 39.3 & 59.0 & 43.5 & 26.0 & 43.5 & 50.1 & 48\\
						baseline + ASFF        & \textbf{40.6} & 59.8 & 45.5 & 27.5 & 45.8 & 51.0 &46\\\hline
					\end{tabular}
		\end{threeparttable}}}
	\end{center}
	\caption{Comparison of ASFF and other fusion operations. APs (\%) are reported on COCO \emph{val}-2017.}
	\label{table:asff}
\end{table}

\begin{figure*}[thbp]
	\begin{center}
		\includegraphics[width=0.81\linewidth]{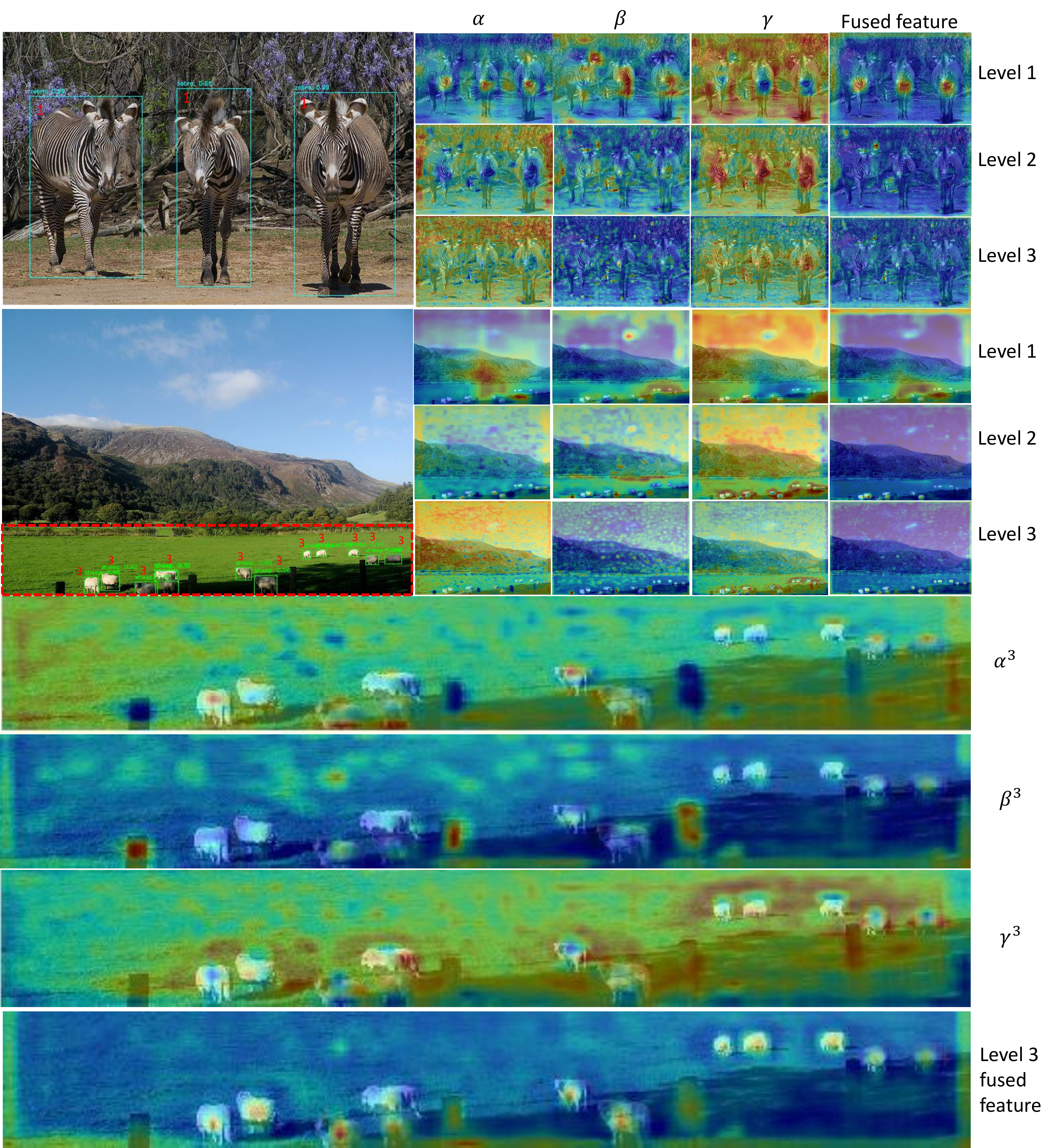}
	\end{center}
	\caption{Visualization of detection results on COCO \emph{val}-2017 as well as the learned weight scalar maps at each level. We zoom in the heat maps of level 3 within the red box for better visualization.}
	\label{fig:weight1}
\end{figure*}

\begin{figure*}[thbp]
	\begin{center}
		\includegraphics[width=0.81\linewidth]{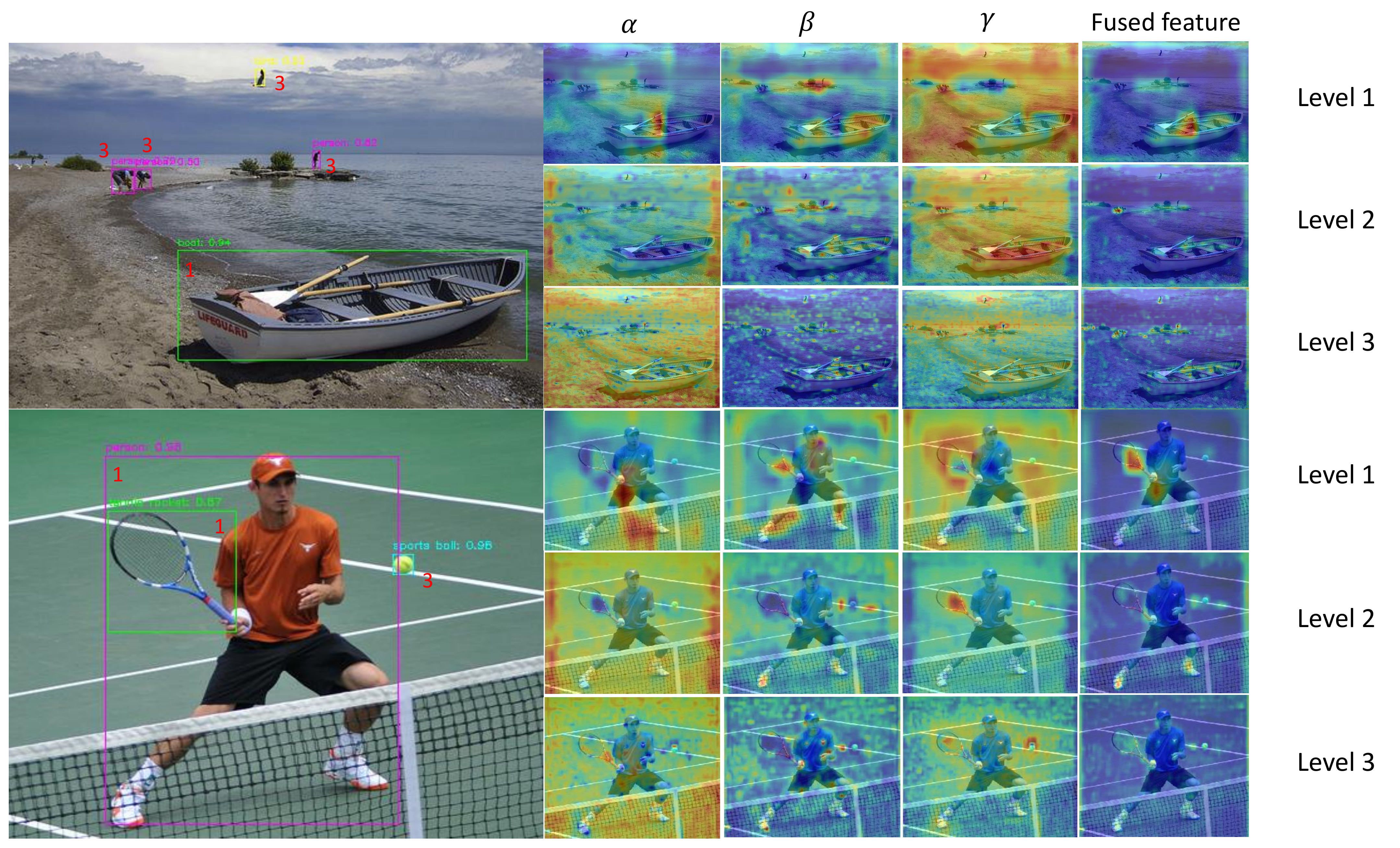}
	\end{center}
	\caption{More qualitative examples when one image has several objects with different sizes.}
	\label{fig:weight2}
\end{figure*}

\begin{table*}[htbp]
	\begin{center}
		\scalebox{0.80}{
			\resizebox{\textwidth}{!}{
				\begin{threeparttable}
					\begin{tabular}{l|l|c|ccc|ccc}
						\Xhline{1.1pt}
						\multirow{2}{*}{Method} & \multicolumn{1}{c|}{\multirow{2}{*}{Backbone}}  & \multicolumn{1}{c|}{\multirow{2}{*}{FPS}} & \multicolumn{3}{c|}{Avg. Precision, IoU:} & \multicolumn{3}{c}{Avg. Precision, Area:} \\
						& \multicolumn{1}{c|}{}                           & \multicolumn{1}{c|}{}                      & 0.5:0.95        & 0.5        & 0.75       & S            & M            & L           \\ \Xhline{1.1pt}
						\emph{two-stages:} &                     &     & & & & & & \\
						Faster w FPN \cite{FPN}            & ResNet-101-FPN                                 & 11.0 (V100)\                                      & 39.8            & 61.3       & 43.3       & 22.9        & 43.3         & 52.6       \\
						Mask R-CNN \cite{mask-rcnn}               & ResNext-101-FPN                           & 6.5 (V100)\                                      & 41.4            & 63.4       & 45.2       & 24.5        & 44.9        & 51.8        \\
						Cascade R-CNN \cite{cascade}     & ResNet-101-FPN            & 9.6 (V100)  \
						& 42.8            & 62.1       &46.3        &23.7         & 45.5         &55.2        \\
						SNIPER \cite{sniper}              & ResNet-101                       &  5.0 (V100)  \
						&46.1             & 67.0       &51.6        &29.6         & 48.9         &58.1       \\ \hline
						\hline
						\emph{one-stages:} &                     &     & & & & & & \\
						SSD300* \cite{ssd}                 & VGG                                            & 43 (Titan X)                                      & 25.1            & 43.1       & 25.8       & --           & --           & --          \\
						RefineDet320 \cite{refinedet}      & VGG                                           & 38.7 (Titan X)
						& 29.4            &49.2        & 31.3       &10.0          & 32.0         &44.4          \\
						RFB Net300 \cite{RFB}                          & VGG                                           & 66.0 (Titan X)\
						& 30.3            & 49.3      & 31.8       & 11.8         & 31.9         & 45.9        \\
						YOLOv3 @320 \cite{yolov3}                  & Darknet-53                                     & 69 (V100)                                       & 28.2            & --        & --       & --         & --         & --        \\
						\textbf{YOLOv3 @320 + ASFF (Ours)}                   & Darknet-53                      & \textbf{63} (V100)                                       & \textbf{36.7}            & \textbf{57.2}        & \textbf{39.5}       & \textbf{15.8}        & \textbf{39.9}         & \textbf{51.3}       \\ 
						
						\textbf{YOLOv3 @320 + ASFF* (Ours)}                   & Darknet-53                      & \textbf{60} (V100)                                       & \textbf{38.1}            & \textbf{57.4}        & \textbf{42.1}       & \textbf{16.1}        & \textbf{41.6}         & \textbf{53.6}       \\ \hline
						SSD512* \cite{ssd}                 & VGG                                            & 22 (Titan X)                                      & 28.8            & 48.5       & 30.3       & --           & --           & --          \\
						RefineDet512\cite{refinedet}      & VGG                                           & 22.3(Titan X)
						& 33.0            & 54.5        & 35.5      & 16.3          & 36.3         &44.3          \\
						RFB Net512 \cite{RFB}              & VGG                                              & 33 (Titan X)                             & 33.8  & 54.2       & 35.9       & 16.2         & 37.1         & 47.4        \\
						RetinaNet500 \cite{focal-loss}            & ResNet-101-FPN                            & 17.8 (V100)                                      & 34.4            & 53.1       & 36.8       & 14.7         & 38.5         & 49.1        \\
						CornerNet-511 \cite{cornernet}     & Hourglass-104                                    & 5.0 (Titan X)
						& 40.5            &56.5        &43.1        &19.4          &42.7          &53.9         \\
						CenterNet-DLA511 \cite{objects}       & DLA-34                                           & 28 (Titan Xp)
						&39.2             &57.1        &42.8        &19.9          &43.0          &51.4         \\
						YOLOv3 @416 \cite{yolov3}                  & Darknet-53                                     & 60 (V100)\                                       & 31.0            & --        & --       & --         & --         & --        \\
						\textbf{YOLOv3 @416 + ASFF (Ours)}                  & Darknet-53                 & \textbf{56} (V100)\                       & \textbf{39.0} & \textbf{60.2} &\textbf{42.5}  & \textbf{19.6}  & \textbf{42.3} & \textbf{51.4}  \\
						\textbf{YOLOv3 @416 + ASFF* (Ours)}                  & Darknet-53                 & \textbf{54} (V100)\                       & \textbf{40.6} & \textbf{60.6} &\textbf{45.1}  & \textbf{20.3}  & \textbf{44.2} & \textbf{54.1}  \\ \hline
						RetinaNet800 \cite{focal-loss}            & ResNet-101-FPN                            & 9.3 (V100)\                                      & 39.1 & 59.1       & 42.3       & 21.8         & 42.7         & 50.2        \\
						FCOS-800 \cite{fcos}                          & ResNet-101-FPN                            & 13.5 (V100)\
						& 41.0            &60.7          &44.1        &24.0          &44.1         &51.0         \\
						NAS-FPN @640 \cite{nas-fpn}              &ResNet-50                                    & 17.8 (P100)\
						& 39.9            & --          & --          & --         &  --         &   --      \\
						NAS-FPN @1280\cite{nas-fpn}              &ResNet-50                                    & 5.2 (P100)\
						& \textbf{46.6}            &  --         &  --         &  --          &--           & --       \\
						YOLOv3 @608 \cite{yolov3}                  & Darknet-53                                     & 52 (V100)\                                       & 33.0            & 57.9        & 34.4       & 18.3         & 35.4        & 41.9        \\
						\textbf{YOLOv3 @608 + ASFF (Ours) } & Darknet-53                                     & \textbf{46.6} (V100)\                                       & \textbf{40.7}  & \textbf{62.9} & \textbf{44.1} & \textbf{24.5} & \textbf{43.6} &\textbf{49.3}   \\
						\textbf{YOLOv3 @608 + ASFF* (Ours) } & Darknet-53                                     & \textbf{45.5} (V100)\                                       & \textbf{42.4}  & \textbf{63.0} & \textbf{47.4} & \textbf{25.5} & \textbf{45.7} &\textbf{52.3}      \\
						\textbf{YOLOv3 @800 + ASFF* (Ours) } & Darknet-53                                     & \textbf{29.4} (V100)\                                       & \textbf{43.9}  & \textbf{64.1} & \textbf{49.2} & \textbf{27.0} & \textbf{46.6} &\textbf{53.4}      \\
						\Xhline{1.1pt}		
					\end{tabular}
		\end{threeparttable}}}
	\end{center}
	\caption{Detection performance in terms of AP (\%) and FPS on COCO \emph{test-dev}.}
	\label{table:coco}
\end{table*}
We perform all the experiments on the bounding box detection track of the challenging MS COCO 2017 benchmark \cite{ms-coco}. We follow the common practice \cite{guide,yolov3} and use the COCO \emph{train-2017} split (consisting of 115k images) for training. We conduct ablation and sensitivity studies according to the evaluation on the \emph{val-2017} split (5k images). For our main results, we report COCO AP on the \emph{test-dev} split (20k images), which has no public labels and requires uploading detection results to the evaluation server.

\subsection{Ablation Study}
\label{exp:ablation}

\paragraph{Solid Baseline.} We first evaluate the contribution of several elements to our baseline detector for better reference. Results are reported in Table \ref{table:baseline}, where BoF denotes all the training tricks mentioned in \cite{bag}, GA denotes the guided anchoring strategy \cite{guide}, and IoU is the additional IoU loss \cite{iouloss} in bounding box regression. From Table \ref{table:baseline}, we can see that all the techniques contribute to accuracy gain, and thanks to them, we deliver a final baseline which reaches an AP of 38.8\%. It is worth to note that the improvement of almost all the components is cost free, as BoF and IoU do not add any additional computation and GA introduces only two $1\times1$ convolution layers for each level of feature maps. The final baseline achieves the speed of 50 FPS on a single Graphics Card of NVIDIA Tesla V100.

\paragraph{Effectiveness of Adjacent Ignore Regions.} To avoid gradient inconsistency, some works \cite{guide,FSAF} ignore the corresponding areas on the two adjacent levels of the chosen level for each target, and the ignored area is the same size as that of the positive one in the chosen level. In YOLOv3, only the center location of the chosen area is positive, and we thus ignore the corresponding center location at the two adjacent levels to follow the ignoring rule. Besides, we denote $\epsilon_{ignore}$ as the ratio of the widths and lengths of the ignored area to that of the target object area, and carry out some experiments with different values of $\epsilon_{ignore}$ to show the effectiveness of the ignoring strategy. Table \ref{table:ignore} reports the study results. We can see that the larger ignore area indeed hurt the performance of the detector, by bringing more false positives.    

\paragraph{Adaptively Spatial Feature Fusion.} ASFF significantly improves the box AP from 38.8\% to 40.6\% as shown in Table \ref{table:asff}. To be more specific, most of the improvements come from $AP_S$ and $AP_M$, yielding increases of 2.9\% and 2.9\% respectively compared with corresponding reference scores. It validates that the representation of high-resolution features is largely improved by the proposed adaptively fusion strategy. Moreover, ASFF only incurs 2 ms additional inference time, keeping the detector run efficiently with 46 FPS.

As described in Sec. \ref{ASFF}, to adaptively fuse the features for each scale, the features at other levels are firstly resized to the same shape before fusion. To make fair comparison, we further report the accuracies of another two common fusion operations (\emph{i.e.} element-wise sum and concatenation) with resized features in Table \ref{table:asff}. We can see in the table that, these two operations improve the accuracy on $AP_S$ and $AP_M$ as ASFF does, but they both sharply downgrade the performance on $AP_L$. These results indicate that the inconsistency across different levels in feature pyramids brings negative influence on the training process and thus leaves the potential of pyramidal feature representation from being fully exploited.

\subsection{Visual Analysis}

In order to understand how the features are adaptively fused, we visualize some qualitative results in Figure \ref{fig:weight1} and \ref{fig:weight2}. The detection results are in the left column. The heat maps of the learned weight scalars and the fused feature activation maps at each level are in the right column. For fused feature maps, we sum up the values among all the channels to visualize the activation maps. The red numbers near the boxes indicate the fused feature level that detects the object. Note that the actual resolutions of the three levels are different, and we resize them to a uniform size for better visualization.

Specifically, in Figure \ref{fig:weight1}, we investigate how ASFF works when all objects in the image have roughly the same size. It is also worth to note that YOLOv3 only takes the center point of the object in the corresponding feature maps as a positive. For the image in the first row, all the three zebras are predicted from the fused feature maps of level 1. It indicates that their center areas are dominated by the original features of level 1 and the resized features within those areas from level 2 and 3 are filtered out. This filtering guarantees that the features of these three zebras at level 2 and 3 are treated as background and do not receive positive gradients in training. Regarding ASFF, in the fusion process of level 2 and 3, the central areas at the resized features from level 1 are also filtered out, and the original features of level 1 will receive no negative gradients in training. For the image in the second row, all the sheeps are predicted by the fused feature maps of level 3. We zoom in the heat maps of level 3 within the red box for better visualization. In fusion, the features from level 1 are kept in the object areas as they contain stronger semantic information, and the features from level 3 are extracted around each object since they are more sensitive for localization.

In Figure \ref{fig:weight2}, we exhibit the images that have several objects of different sizes. Most of the fusion cases at the corresponding level are similar to the ones in Figure \ref{fig:weight1}. Meanwhile, one may notice that the tennis racket in the second image is predicted from level 1, but the heat maps show that the main features within its central area are taken from the resized feature of level 2. We speculate that although the tennis racket is predicted from level 1 due to heuristic size selection, the features from level 2 are more discriminative in detecting it since they contain richer clues of lines and shapes. Thanks to our ASFF module, the final feature can be adaptively learned from optimal fusion, which contributes in particular to detecting challenging objects. Please see more visual results in the supplementary material.

\subsection{Evaluation on Other Single-Shot Detectors}
To better evaluate the performance of the proposed approach, we carry out additional experiments with another representative single-shot detector, namely RetinaNet \cite{focal-loss}. First, we directly adopt the official implementation \cite{maskrcnnbenchmark} to reproduce the baseline. We then add ASFF behind the pyramid feature maps from P3 to P5 on FPN, similar to Figure \ref{fig:frame}. As shown in Table \ref{table:retina},  
ASFF consistently increases the accuracy of RetinaNet with different backbones (\emph{i.e.} ResNet-50 and ResNet-101).

\begin{table}[htbp]
	\begin{center}
		\scalebox{.47}{
			\resizebox{\textwidth}{!}{
				\begin{threeparttable}
					\begin{tabular}{l|l|lll}
						\hline
						Method                                                                    & Backbone      & AP   & AP50 & AP70 \\ \hline
						\multirow{4}{*}{\begin{tabular}[c]{@{}l@{}}RetinaNet \cite{focal-loss}\\ @800\end{tabular}} & R50-FPN       & 35.9 & 55.4 & 38.8 \\
						& R50-FPN+ASFF  & \textbf{37.4} & \textbf{56.5} & \textbf{39.9} \\
						& R101-FPN      & 39.1 & 59.1 & 42.3 \\
						& R101-FPN+ASFF & \textbf{40.1} & \textbf{59.3} & \textbf{42.8} \\ \hline
					\end{tabular}
		\end{threeparttable}}}
	\end{center}
	\caption{Contribution of ASFF to RetinaNet. APs (\%) are reported on COCO \emph{val}-2017.}
	\label{table:retina}
	\vspace{-.4cm}
\end{table}

\subsection{Comparison to State of the Art}

We evaluate our detector on the COCO \emph{test-dev} split to compare with recent state-of-the-art methods in Table \ref{table:coco}. Our final model is YOLOv3 with ASFF*, which is an enhanced ASFFversion by integrating other lightweight modules (\emph{i.e.} DropBlock \cite{dropblock} and RFB \cite{RFB}) with 1.5$\times$ longer training time than the models in Section \ref{exp:ablation}. 
Keeping the high efficiency of YOLOv3, we successfully uplift its performance to the same level as the state-of-the-art single-shot detectors (\emph{e.g.}, FCOS \cite{fcos}, CenterNet \cite{objects}, and NAS-FPN \cite{nas-fpn}), as shown in Figure \ref{fig:speed}. Note that YOLOv3 can be evaluated at different input resolutions with the same weights, and when we lower the resolution of input images to pursue much faster detector, ASFF improves the performance more significantly. 

\section{Conclusion}

This work identifies the inconsistency across different feature scales as a primary limitation for single-shot detectors with feature pyramids. To address this, we propose a novel ASFF strategy which learns the adaptive spatial fusion weight to filter out the inconsistency during training. It significantly improves strong baselines with tiny inference overhead and achieves a state-of-the-art speed and accuracy trade-off among all single-shot detectors.

{\small
	\bibliographystyle{ieee_fullname}
	\bibliography{cvpr}
}

\end{document}